\documentclass[11pt]{article}
\usepackage{amsfonts, amssymb, amsmath}
\usepackage{algorithm, algorithmic, amsthm}
\usepackage{graphicx}
\usepackage{color}
\usepackage{hyperref}
\usepackage[top=2.5cm, bottom=3cm, left=3cm, right=3cm]{geometry}

\usepackage{subfigure}

\usepackage{hyperref}
\usepackage{url}
\usepackage{graphicx, wrapfig}
\usepackage{amsfonts, amssymb, amsmath}
\usepackage{algorithm, algorithmic, amsthm}

{%
\begin{list}{#1}{
\vspace{-\topsep}
\vspace{-\partopsep}
\setlength{\itemindent}{0cm}
\setlength{\rightmargin}{0cm}
\setlength{\listparindent}{0cm}
\settowidth{\labelwidth}{#1}
\setlength{\leftmargin}{\labelwidth}
\addtolength{\leftmargin}{\labelsep}
\setlength{\itemsep}{0cm}
}%
}%
{%
\end{list}
\vspace{-\topsep}
\vspace{-\partopsep}
}

{\begin{enumerate}%
}%
{\end{enumerate}}

\theoremstyle{plain}

\newtheorem*{lemma*}{Lemma}

\newtheorem*{prop*}{Proposition}

\theoremstyle{definition}

\newtheorem*{defn*}{Definition}

\newtheorem*{exmp*}{Example}

\newtheorem*{conj*}{Conjecture}

\theoremstyle{remark}

\newtheorem*{rmk*}{Remark}

\title{\sc Event Identification as a Decision Process with Non-linear Representation of Text}
\author{\sf YukunYan$^1$\thanks{The work is done when the first author worked as intern at DeeplyCurious.AI.} \ \;Daqi Zheng$^2$\; Zhengdong Lu$^2$ \; Sen Song$^{1}$ \\ $^1$Dept. of Bio-medical Engineering \\ Tsinghua University\\ {\tt \{yanyk13, songsen\}@mails.tsinghua.edu.cn}\\ $^2$DeeplyCurious.AI\\ {\tt \{da, luz\}@deeplycurious.ai} }

\date{}
\begin{document}
\maketitle

\begin{abstract}
 We propose \textbf{scale-free Identifier Network(sfIN)}, a novel model for event identification in documents. In general, sfIN first encodes a document into multi-scale memory stacks, then extracts special events via conducting multi-scale actions, which can be considered as a special type of sequence labelling. The design of large scale actions makes it more efficient processing a long document. The whole model is trained with both supervised learning and reinforcement learning. 
\end{abstract}

\section{Introduction}
Specific information extraction is a basic stage of text understanding, which is always conducted as a sequence labelling task. Typical approaches include traditional linear models and recurrent neural network based model. The previous type of models, Hidden Markov Models (HMM) and Conditional Random Fields (CRF) included, are based on hand-craft features, which lacks information such like the order of words. On the other side, neural based models\cite{hochreiter1997long}\cite{cho2014properties} achieved outstanding successes recently. The most significant difference between two types of models is neural based models use distributed presentation and non-linear transformation, which enable the algorithms to build a more complex language model. The most successful models are LSTM-crf\cite{NER_bilstm_crf}\cite{huang2015bidirectional} and LSTM-CNNs-crf\cite{ma2016end}. 

One research\cite{neural} about the human brain cortex shows that, several parts of cortex are activated listening speeches, corresponding to different levels of language structure. Moreover, this phenomenon only happens when the listener knows the language, which infers using a multi-level distributed representation maybe a providential way to build a language model\cite{collobert2011natural}. Many attempts were made based on multi-scale recurrent neural network, which can be divided into two types. The first type has several recurrent layers and each layer has its own update period\cite{el1996hierarchical}, CW-RNN\cite{koutnik2014clockwork} included. While the second type\cite{bahdanau2016end}\cite{chung2015gated}\cite{chung2016hierarchical} uses a gating mechanism controlling the flow from low level hidden state to high level ones, such like hierarchical recurrent neural networks. Here, sfIN is designed in a different way. Its representation levels correspond to the language structures, including word, sentence and paragraph.

\section{Architecture}
Different from models like LSTM-crf and other neural network based models, mRR encode text into a hierarchical memory stack, which enable more complex non-linear transformation of the whole text. After establishing the memory stack, a RNN based controller will read part of the memory at each time and make an action to predict current tag. There are three read-heads which will be updated after an action is made, indicating which part of memory is accessible. The whole process will end when one of the read-head reaches the bottom of text. \textsl{Figure 1} shows the architecture of mRR.
\begin{figure}[h]
	\centering
	\includegraphics[width=0.5\textwidth]{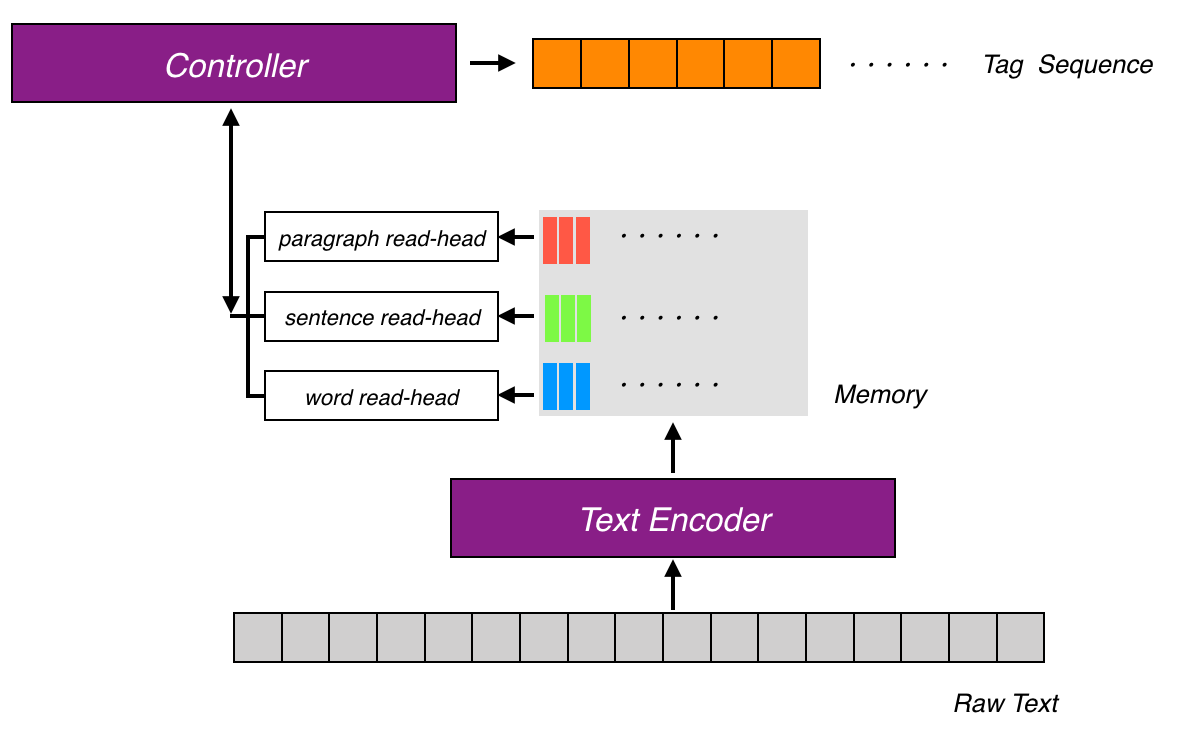}
	\caption{Coarse-Reader Network}
\end{figure}

\textbf{Road Map} The remainder of this section is organized as follows. Section 2.1 describes  \textbf{Text Encoder} part of mRR and Section 2.2 shows the \textbf{Controller} part. 

\subsection{Text Encoder}
Text encoder takes not only the raw text as input but also the structure information, and output a hierarchical memory, which has three level parts: word level, sentence level and paragraph level.
A memory is generally defined as a matrix with potentially infinite size, while here we limit the memory with three pre-claimed matrix $N_{w} \times d_{w}$, $N_{s} \times d_{s}$, $N_{p} \times d_{p}$, with $N_{*}$ locations and $d_{*}$ values in each location at each level. $N_{*}$ is always instance-dependent and is pre-determined by the algorithm. In our implementation, memory of different level has different $d_{*}$.
$$M_{text} = f_{te}(X, IX_{s}, IX_{p}) \eqno{(1)}$$
$$M_{text} = [M_{w}, M_{s}, M_{p}] \eqno{(2)}$$
with $X = [x_1, x_2, ..., x_{N_{w}}]$, $IX_{s} = [ix_{s_{1}}, ix_{s_{2}}, ... ix_{N_{s}}]$ denotes the begin indexes of sentences, $IX_{p} = [ix_{p_{1}}, ix_{p_{2}}, ... ix_{N_{p}}]$ denotes the begin indexes of paragraphs.
The Text Encoder have three part: word encoder, sentence encoder and paragraph encoder
\begin{figure}[h]
	\centering
	\includegraphics[width=0.5\textwidth]{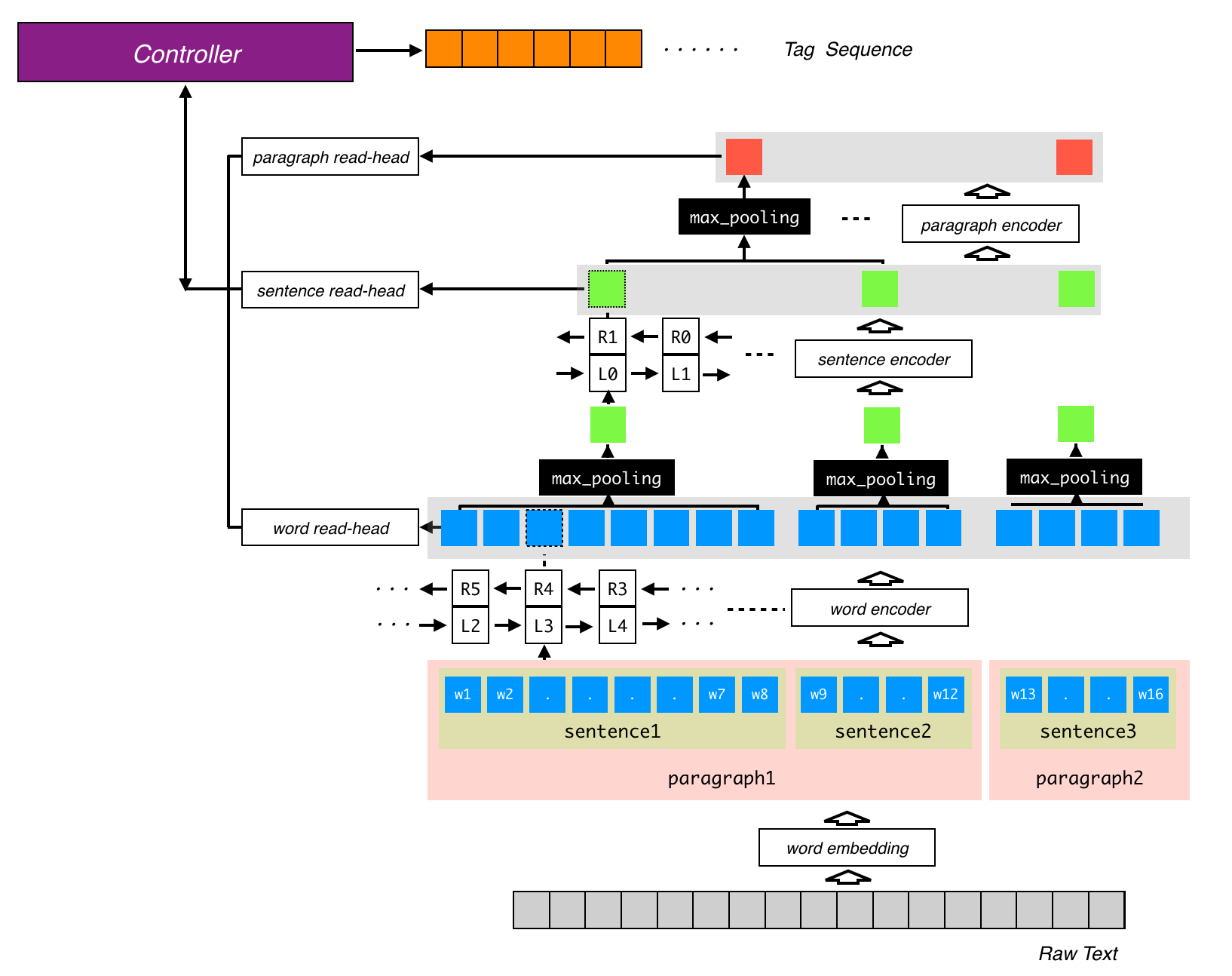}
	\caption{Text Encoder}
\end{figure}

\subsubsection{word encoder}
The word encoder takes raw text as input and output the word level memory $M_{w} \in N_{w} \times d_{w}$, where $N_{w}$ is the number of words in the document, $d_{w}$ is the dimension of word level memory. As illustrated in \textsl{Figure 3}, the establishing process is as follow: 1. We apply a word embedding layer on the raw text to gain several vector sequences, each corresponding to a sentence. 2. The vector sequences was put into a bidirectional Long short term memory (bi-LSTM) layer to generate the memory matrix $M_{w}$ by concatenating all hidden states.

\begin{figure}[h]
	\centering
	\includegraphics[width=0.5\textwidth]{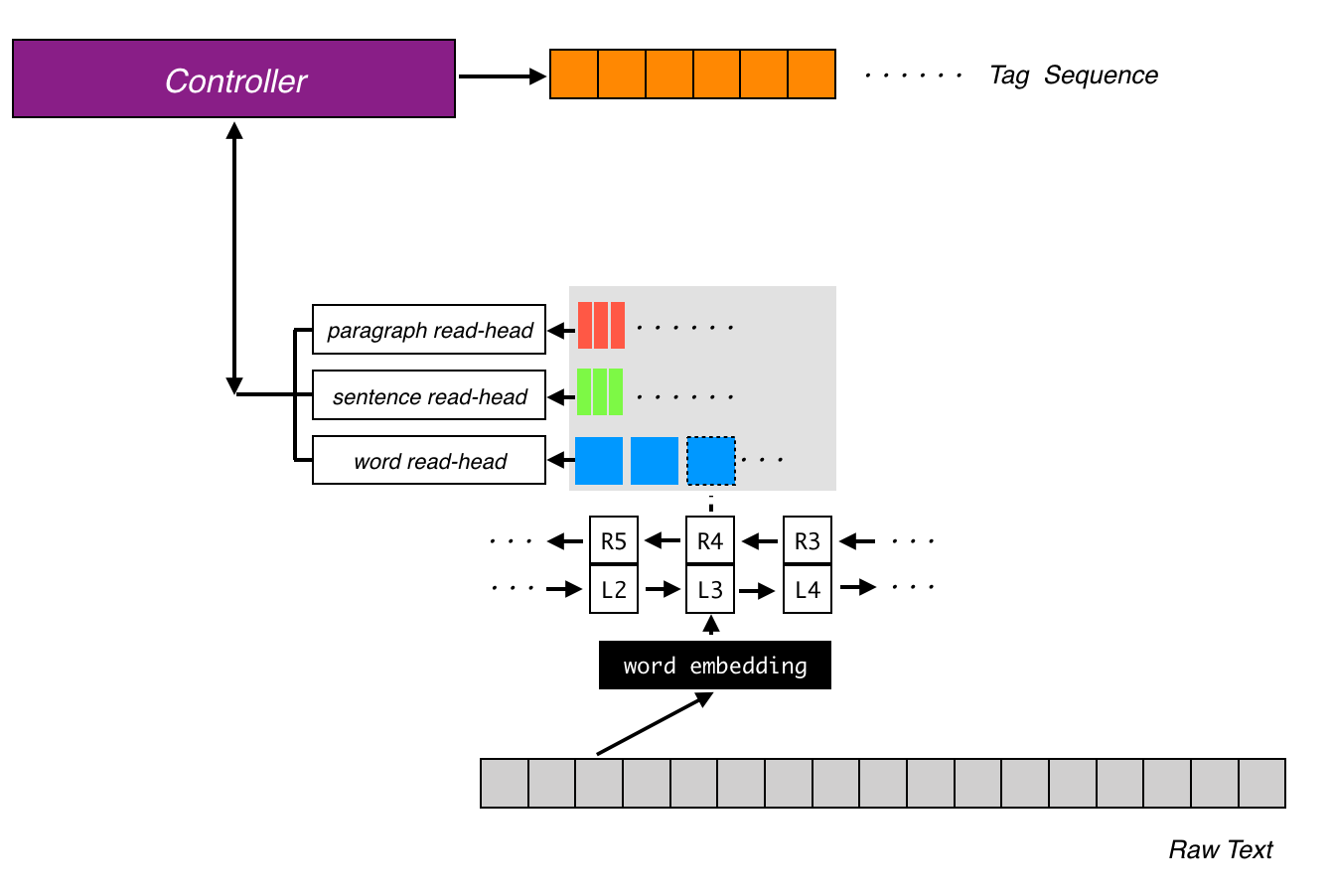}
	\caption{establishment of word level memory}
\end{figure}
\subsubsection{sentence encoder}
Inspired by convolutional neural networks, we apply element-wise max-pooling on the word level memory to generate 'sentence vector', in extracting global feature from local features. Another bi-LSTM layer is used to generate sentence level memory matrix $M_{s}$ like generating $M_{w}$, as illustrated in \textsl{Figure 4}.
\begin{figure}[h]
	\centering
	\includegraphics[width=0.5\textwidth]{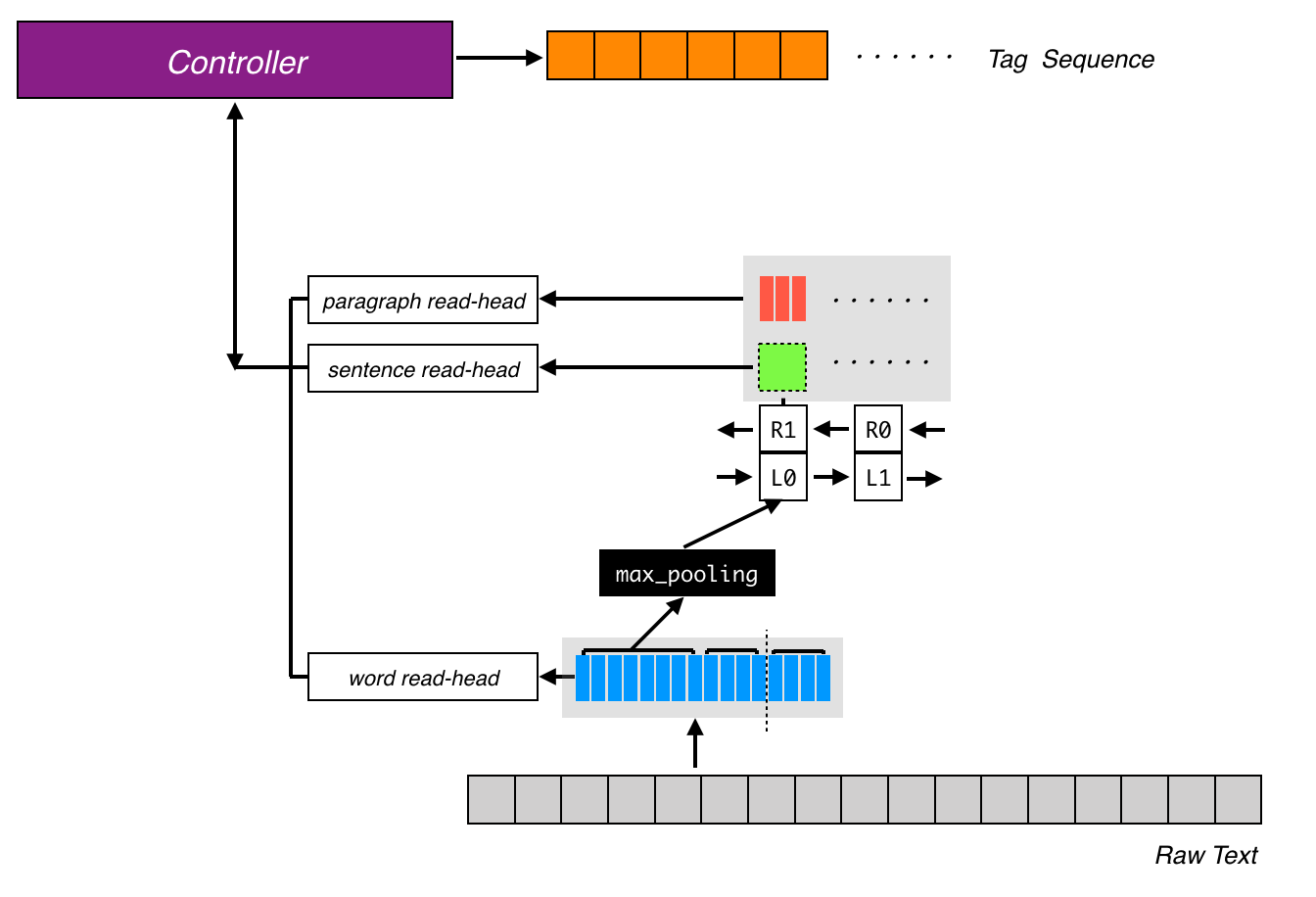}
	\caption{establishment of sentence level memory}
\end{figure}
\subsubsection{paragraph encoder}
For each paragraph, we apply element-wise max-pooling on the sub matrix of $M_{s}$ corresponding to the sentences belong to it and generate $M_{p}$, as shown in \textsl{Figure 5}
$$M_{p} = max_{element}([v_{s_1}, v_{s_2} ... v_{s_n}]) \eqno{(9)}$$
where $s_1, s_2, ..., s_n \in p$
\begin{figure}[h]
	\centering
	\includegraphics[width=0.5\textwidth]{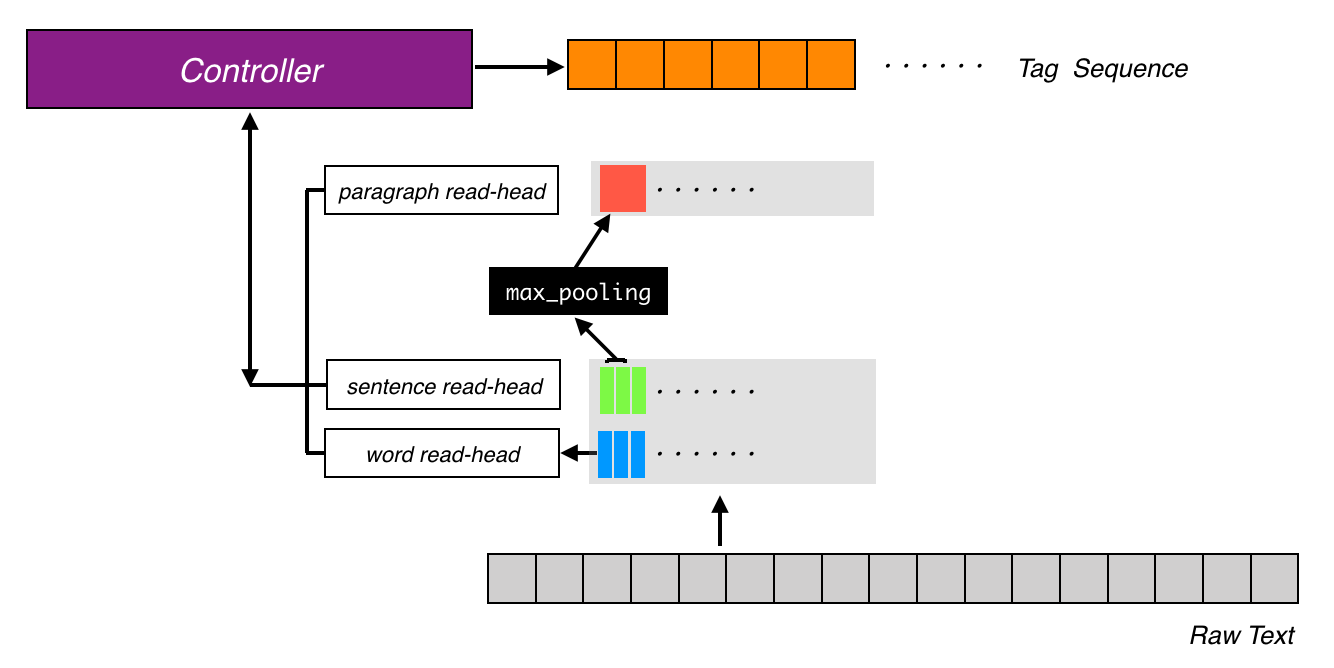}
	\caption{establishment of paragraph level memory}
\end{figure}

\subsection{Controller}
The structure of the controller part is a RNN layer and nine feed-forward neural networks (FNN), which has an output of one dimension. At each time, the RNN uses three read-heads to read the hierarchical memory and update its hidden state, which is feed to the FNN to generate an action. The tag sequence is add to previous result and the location of read-heads is updated at the same time, as illustrated in \textsl{Figure 6}.

The controller is trained as an action agent, which can read part of the hierarchical memory and make a choice of actions (and generate part of tag sequence) at each time step. There are nine available actions corresponding to nine feed-forward neural networks as follow: mark a word/sentence/paragraph as non event, current event or new event. Available actions and corresponding tag sequence are shown in \textsl{Table 1}
\newline

\begin{figure}[h!]
	\centering
	\includegraphics[width=0.5\textwidth]{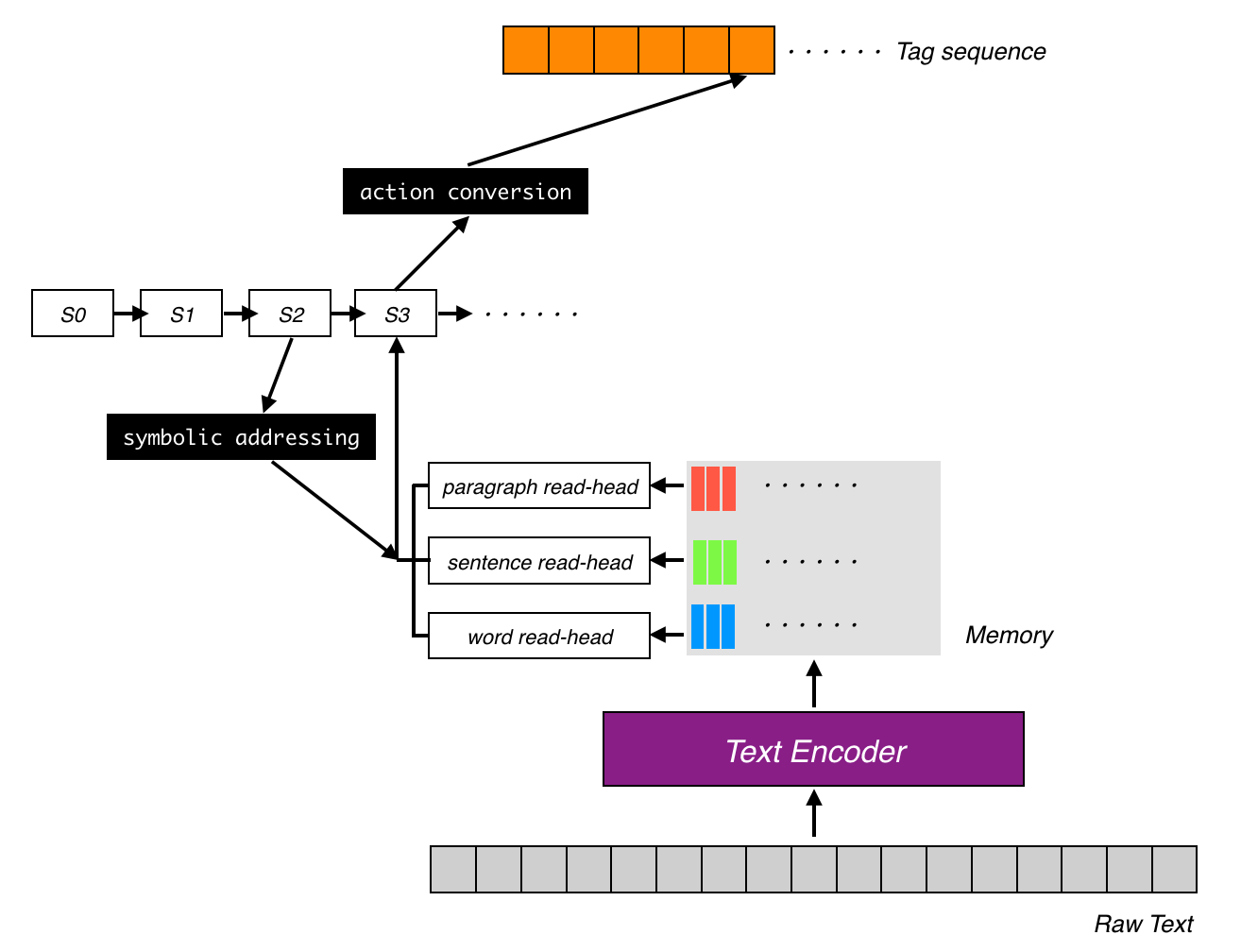}
	\caption{controller}
\end{figure}

\begin{table}[H] \footnotesize
	\begin{center}
		\begin{tabular}{|l|l|l|l|}
			\hline
			action level & non-event  & current event &  new event\\
			\hline
			word         & -1 & mark & mark + 1     \\
			\hline
			sentence     & $\overbrace{[-1, .., -1]}^{s_{len}}$  &  $\overbrace{[mark, .., mark]}^{s_{len}}$  &  $\overbrace{[mark+1, ... mark + 1]}^{s_{len}}$     \\
			\hline
			paragraph    & $\overbrace{[-1, .., -1]}^{p_{len}}$  &  $\overbrace{[mark, .., mark]}^{p_{len}}$ & $\overbrace{[mark+1, .., mark+1]}^{p_{len}}$   \\
			\hline
		\end{tabular}
	\end{center}
	\caption{available actions and corresponding tag sequence}
	\label{t:results}
\end{table}%

\textsl{Read Memory}: The decision process is conducted from the beginning of text to the end. There is a vector indicating the current location, which has three dimensions corresponding to three level of the accessible hierarchical memory, initialized as $[0, 0, 0]$. At each time step, the available memory is loaded in to the Controller, as illustrated in \textsl{Figure 6}\\
\newline

\begin{figure}[h]
	\centering
	\includegraphics[width=0.5\textwidth]{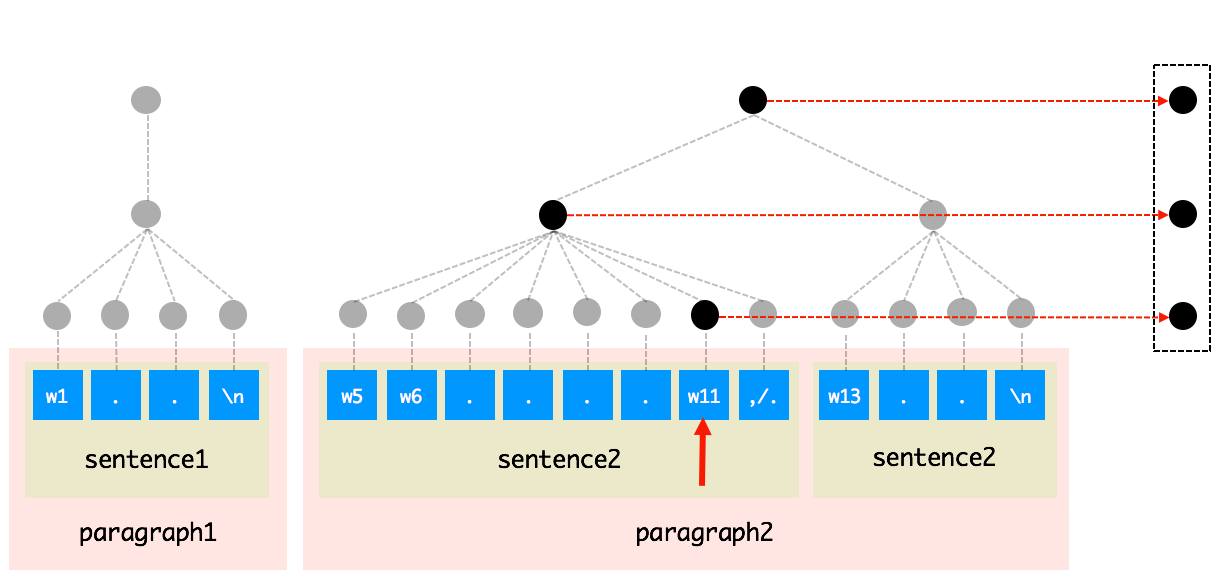}
	\caption{reading memory, current location vector is [11, 2, 2]}
\end{figure}

\textsl{Generate Action}: The chosen part of memory is used to update the state of controller, which then generates nine scores. The action with the highest score is chosen to generate tag sequence along \textsl{table 1} at each time step. \\
$$a_{i} = \arg max([s_1, s_2, .. s_9])$$
\newline
\textsl{Update Location}: The action is also used to update location vector mentioned above based on which level action was made. The new vector points at the next word if the action is at word level or the first word of the next section if the action is at sentence/paragraph level. An instance is shown in \textsl{Figure 8} \\ 

\begin{figure}[h]
	\centering
	\includegraphics[width=0.5\textwidth]{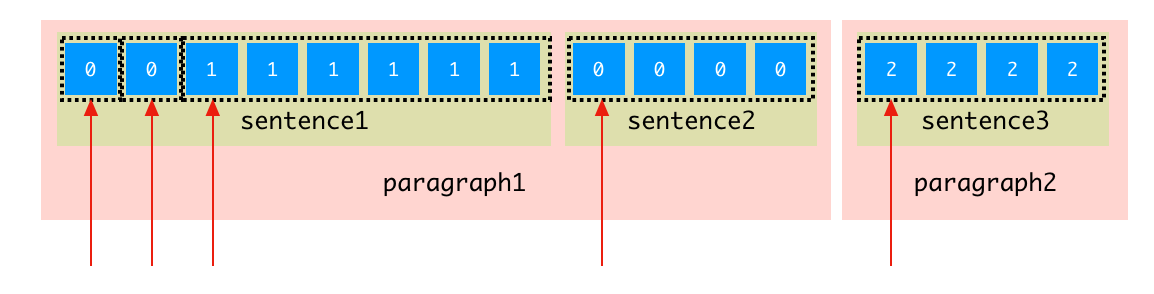}
	\caption{action sequence: [word-none, word-none, sentence-new, sentence-none, paragraph-new]}
\end{figure}

\section{Dataset}
We used a law papers dataset, each contains the information of a criminal and his criminal record. In each sample, there can be one or several stealing events with different time, location and victims. The length of a sample ranges from 1500 words to 7000 words, and the number of event in each one ranges from 1 to 74. The whole dataset has 8299 samples and labelled by 5 individuals.

\section{Training}
We use both supervised learning and reinforcement learning (policy gradient) strategy in training our model. At each time step, we generate action labels by comparison the predicted tag sequence and ground true. The action label will be set as '1' only if all of its predictions are right, otherwise, '0'. However, the may be several actions are all correct. So, we use multinomial sampling of the correct actions at each time step to guarantee it follows the right path. Because the length of text influences the speech of applying, we want the model can take actions that corresponding to longer sequence as long as it is correct. This can not be achieved by supervised learning, because the fewest actions policy may not gain the highest accuracy. So we introduced a policy gradient based method that after processing one sample, the model will gain a reward based on the number of actions divided by the length of text. In this way, the model finds a balance between efficiency and precise.

\section{Experiments}
We compared our model and LSTM-crf, finding that the f1 value of mRR reaches 93.02\%, which exceeds about 10\%.
\begin{table}[H] \footnotesize
	\begin{center}
		\begin{tabular}{|l|l|l|l|}
			\hline
			model       & test accuracy     & test recall     &  test f1  \\
			\hline
			LSTM-crf    & $89.01$\%         & -               & $79.00$\% \\
			\hline
			mRR-ne      & $97.35$\%         & -               & $93.02$\% \\
			\hline
			mRR-le      & $97.08$\%         & -               & $92.09$\% \\
			\hline
		\end{tabular}
	\end{center}
	\caption{available actions and corresponding tag sequence}
	\label{t:results}
\end{table}%

\section{Conclusion}

We have demonstrate that multi-scale language model can combine global features and local features to help extract key information of an ontology, which performances better the single RNN layer models. And the model trained to take large scale actions has great advantage on processing efficiency over previous models.

\section{Discussion}
We found that mRR is perfectly good at tagging tasks on long texts. It prefer low level actions at some key words which infers the function in the section, for instance, 'the People's Court establishes the truth based on facts' and high level actions at some 'no big deal' sections, like the basic information of a criminal. This is very intuitive that the same progress happens when human process the same task.

\newpage
\bibliographystyle{abbrv}
\small{\bibliography{reference}}
\end{document}